\documentclass[lettersize,journal]{IEEEtran}
\usepackage{amsmath,amsfonts}
\usepackage{algorithmic}
\usepackage{array}
\usepackage[caption=false,font=normalsize,labelfont=sf,textfont=sf]{subfig}
\usepackage{textcomp}
\usepackage{stfloats}
\usepackage{url}
\usepackage{verbatim}
\usepackage{graphicx}
\usepackage[hidelinks]{hyperref}
\usepackage{cleveref}
\usepackage{balance}
\usepackage{algorithm}
\usepackage{subcaption}
\usepackage{multirow} 
\crefname{section}{Section}{Sections}
\crefname{table}{Table}{Tables}
\crefname{figure}{Fig.}{Figs.}
\usepackage{tikz,xcolor,hyperref}
\definecolor{lime}{HTML}{A6CE39}
\DeclareRobustCommand{\orcidicon}{%
    \begin{tikzpicture}
    \draw[lime, fill=lime] (0,0) 
    circle [radius=0.16] 
    node[white] {{\fontfamily{qag}\selectfont \tiny ID}};    \draw[white, fill=white] (-0.0625,0.095) 
    circle [radius=0.007];    \end{tikzpicture}
    \hspace{-2mm}}
\foreach \x in {A, ..., Z}{%
    \expandafter\xdef\csname orcid\x\endcsname{\noexpand\href{https://orcid.org/\csname orcidauthor\x\endcsname}{\noexpand\orcidicon}}
    }
    
\begin{document}
\title{Leveraging Unknown Objects to Construct Labeled-Unlabeled Meta-Relationships for Zero-Shot Object Navigation}

\author{Yanwei Zheng\orcidA{}, \IEEEmembership{Member, ~IEEE},
Changrui Li\orcidB{},
Chuanlin Lan\orcidC{},
Yaling Li\orcidD{},
Xiao Zhang\orcidE{}, \IEEEmembership{Member, ~IEEE},
Yifei Zou\orcidF{}, \IEEEmembership{Member, ~IEEE},
Dongxiao yu\IEEEauthorrefmark{1} \orcidG{}, \IEEEmembership{Senior Member, ~IEEE},
Zhipeng Cai\orcidH{}, \IEEEmembership{Fellow, ~IEEE}

\thanks{
\IEEEauthorrefmark{1}Corresponding author.

Yanwei Zheng, Changrui Li, Yaling Li, Xiao Zhang, Yifei Zou, and Dongxiao Yu are with the School
of Computer Science and Technology, Institute of Intelligent Computing,
Shandong University, Qingdao 266237, China (e-mail: zhengyw@sdu.edu.cn; crlee@mail.sdu.edu.cn; yalingli@mail.sdu.edu.cn; xiaozhang@sdu.edu.cn; yfzou@sdu.edu.cn; dxyu@sdu.edu.cn).
 
Chuanlin Lan is a Ph.D. student in the Department of Electrical Engineering at City University of Hong Kong, Hong Kong 999077, China(e-mail: cllan2-c@my.cityu.edu.hk).

Zhipeng Cai is with the Department of Computer Science, Georgia State University, Atlanta, GA 30303, USA (e-mail: zcai@gsu.edu).
}
}

\maketitle

\begin{abstract}
Zero-shot object navigation (ZSON) addresses situation where an agent navigates to an unseen object that does not present in the training set. 
Previous works mainly train agent using seen objects with known labels, and ignore the seen objects without labels. 
In this paper, we introduce seen objects without labels, herein termed as ``unknown objects'', into training procedure to enrich the agent's knowledge base with distinguishable but previously overlooked information.
Furthermore, we propose the label-wise meta-correlation module (LWMCM) to harness relationships among objects with and without labels, and obtain enhanced objects information. Specially, we propose target feature generator (TFG) to generate the features representation of the unlabeled target objects. Subsequently, the unlabeled object identifier (UOI) module assesses whether the unlabeled target object appears in the current observation frame captured by the camera and produces an adapted target features representation specific to the observed context. In meta contrastive feature modifier (MCFM), the target features is modified via approaching the features of objects within the observation frame while distancing itself from features of unobserved objects. Finally, the meta object-graph learner (MOGL) module is utilized to calculate the relationships among objects based on the features. Experiments conducted on AI2THOR and RoboTHOR platforms demonstrate the effectiveness of our proposed method. 
\end{abstract}


\section{Introduction}
\IEEEPARstart{O}{bject} navigation, which can utilize cameras and computer vision technology to detect and navigate within traffic environments, has shown great potential in optimizing intelligent transportation systems and brings benefits to autonomous vehicles due to its ability to provide accurate environmental perception and information processing \cite{qin2021light, gao2018object, guo2021coff}. In the realm of object navigation, where agents are tasked with locating target objects based solely on initial visual cues, conventional methods excel within the confines of their training environments but falter when faced with objects not encountered during their training phase. This gap has led to the development of Zero-shot Object Navigation (ZSON), a paradigm designed to equip agents with the ability to navigate towards unseen object—those absent from the training dataset, in contrast to seen objects that were part of the training phase. To enhance the efficacy of navigating to unseen objects, several innovative approaches have been proposed. Yang et al. \cite{yang2018visual}  leveraged external datasets to imbue agents with prior knowledge of unseen objects, while Khandelwal et al. \cite{khandelwal2022simple} utilized the broad capabilities of the CLIP model \cite{radford2021learning} for direct feature extraction from observations. Furthermore, \cite{zhang2022generative} innovated by generating visual features for unseen objects from their category semantic embeddings, integrating these features seamlessly into the agents' operational environment.

\begin{figure*}[t]
\centering
\includegraphics[width=1\textwidth]{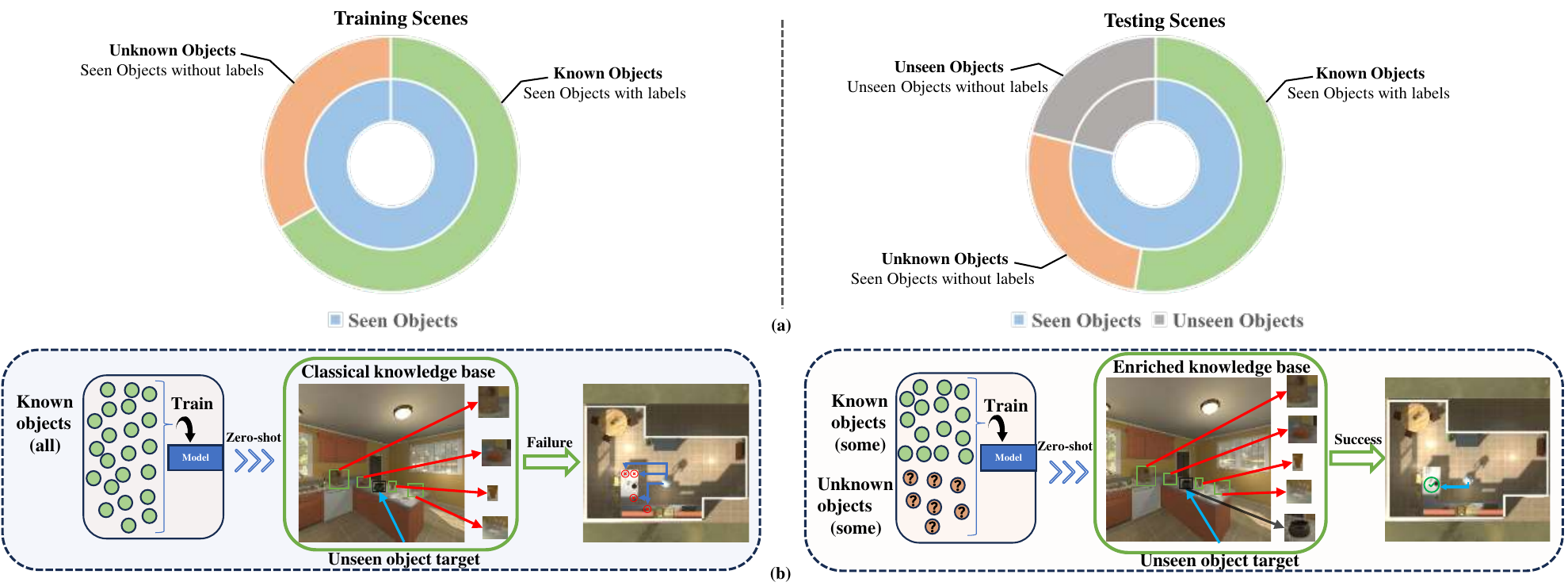}
\caption{(a) Details of split target objects. (b) The scenario on the left is training agent solely on limited known objects information and finally agent navigates to the wrong object. The scenario on the right involves introducing unknown objects into the training phase to provide the agent with more information and explicit guidance.}
\label{fig1}
\end{figure*}

Despite these advancements, a common limitation persists across these methods: the reliance on training with seen objects that are labeled, while neglecting the potential learning from unlabeled seen objects. This oversight can lead to suboptimal decision-making, as agents might incorrectly navigate towards labeled seen objects when attempting to locate unseen targets, as illustrated in the left panel of Fig.\ref{fig1} (b). To mitigate this issue, our approach involves the novel introduction of ``unseen objects'' during the training phase. By deliberately concealing the labels of some objects within the training set, we create a dichotomy of known objects and unknown objects, allowing both categories to coexist within the training scenes, as depicted in Fig.\ref{fig1} (a) and the right panel of Fig.\ref{fig1} (b). In this way, it not only enriches the information available to the agent but also enhances its ability to distinguish between and navigate towards both seen and unseen objects.

Furthermore, we propose the label-wise meta-correlation module (LWMCM) to harness relationships among objects with and without labels, and obtain enhanced objects information. 
Specifically, as shown in Fig.\ref{fig2}, our approach employ object relationships embedding $z_r$ to boost the navigation performance. Initially, we employ a unlabeled target feature generator (TFG, Sec.~\ref{sec:tfg}) to generate the features representation $g_t$ for the target object based on its semantic attributes. Then we introduce an unlabeled object identifier module (UOI, Sec.~\ref{sec:uoi}) to evaluate whether an unlabeled object appears in the frame. If so, the module output an intermediate features denoted as $f_t$ representing the unlabeled target object's features in the observed frame. Next, to establish connections among object features, we employ a meta contrastive feature modifier module (MCFM, Sec.~\ref{sec:mcfm}). Within this module, the intermediate features $f_t$ is brought closer to the features of objects that co-occur with it in the observation frame and is pushed away from the features of objects that are not present in the frame. Following this stage, we employ meta object-graph learner (MOGL, Sec.~\ref{sec:mogl}) to calculate relationships embedding $z_r$, representing the relationships among the objects. Subsequently, the relationships embedding is concatenated with both the observation embedding $z_o$ and the target embedding $z_t$. The combined representation is then input into an LSTM-based policy learner to compute the subsequent action.(Sec.~\ref{sec:np})

Tested in the complex environments of AI2THOR \cite{kolve2017ai2} and RoboTHOR \cite{deitke2020robothor}, our method outperforms existing benchmarks in navigating towards unseen objects in novel scenes. Our main contributions are summarized as follows:

\begin{itemize}

\item We enhance the training process by incorporating unknown objects to enrich the dataset with diverse and informative content.
\item We propose the label-wise meta-correlation module (LWMCM) to harness relationships among objects with and without labels, and obtain enhanced objects information. 
\item Our approach achieves state-of-the-art performance in both the AI2THOR and RoboTHOR simulation environments.
\end{itemize}

The rest of the paper is organized as follows. \cref{related work} reviews relevant research about object navigation. \cref{method} presents the details of our module. ~\cref{experiment} presents the experimental setting, results on two datasets and result analysis. ~\cref{conclusion} concludes the paper.

\section{Related Work} \label{related work}
\subsection{Object Navigation}

Object Navigation (ObjNav) using objects as targets can be divided into seen object navigation and unseen object navigation (zero-shot object navigation, ZSON). A number of approaches have been proposed for solving seen object navigation in recent years, e.g. constructing topological graphs or metric maps methods \cite{chaplot2020neural, savinov2018semi,ramakrishnan2021exploration, gupta2017cognitive}, Transformer \cite{vaswani2017attention} based methods \cite{du2021vtnet, fukushima2022object, fang2019scene}, RL \cite{feng2024security, mnih2016asynchronous} based methods \cite{mirowski2018learning, zhu2017target},  and relational GCN \cite{kipf2016semi} methods \cite{du2020learning,dang2022unbiased, zhang2021hierarchical}. Besides, methods focused on historical state analysis and enhancing process memory have also achieved good performance \cite{dang2023multiple,du2023object, zhang2023layout}.

Our work mainly focuses on unseen objects in the unseen environments. \cite{yang2018visual} firstly proposes navigating to unseen objects and uses external datasets to provide a prior knowledge. \cite{khandelwal2022simple} uses the trained CLIP model \cite{radford2021learning} to directly extract features from observations. However, using external information or large-scale model may contain unseen classes that appear during the training phase. Thus Strict Zero-Shot Object Navigation (S-ZSON) is proposed by \cite{zhao2023zero} which does not allow image and word embedding samples of unseen classes to appear during the training phase. \cite{zhang2022generative} employs a generative meta-adversarial network to generative features of targets only using semantic embedding and trained Resnet18 \cite{he2016deep} model on ImageNet \cite{deng2009imagenet}. We also adopt the latter criterion as the standard and concentrate our attention on incorporating unknown objects to the training process and obtain enhanced objects information.


\subsection{Meta-learning Method}
Drawing inspiration from human learning, meta-learning, or learning to learn, leverages the ability to acquire prior knowledge during training and then rapidly adapts this learned knowledge to novel tasks \cite{thrun1998learning}. The gradient-based meta-learning approach \cite{nichol2018first,hochreiter2001learning, cao2021meta} can be interpreted as the process of learning an effective parameter initialization, allowing the network to perform well with only a few gradient updates. \cite{wortsman2019learning, zhang2022generative} use this method for adapting in new scenes. In our work, we take the gradient-based meta-learning to optimize the network performance based on the specific testing conditions.

\begin{figure*}[!htbp]
\centering
\includegraphics[width=1\textwidth]{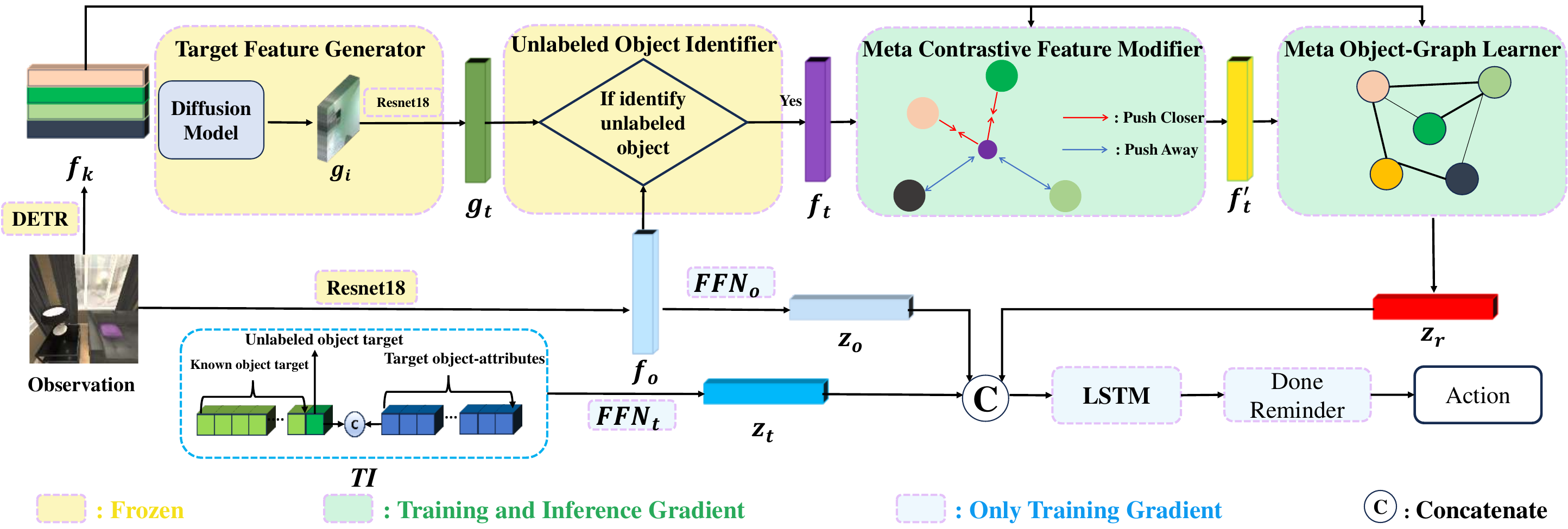}
\caption{\textbf{Model Overview}: 
$TI$: target indicator, $f_o$: observation features, $f_k$: known objects feature from DETR, $g_i$: generative images of unlabeled targets, $g_t$: generative features, $f_t$: intermediate features of UOI, $f_t'$: output of MCFM, $z_o$: observation embedding, $z_t$: target embedding, $z_r$: relationships embedding. Our LWMCM network consists of four parts to get object features: TFG, UOI, MCFM and MOGL. Then the joint features of target embedding, observation embedding and relationships embedding are input into an LSTM network to predict the next action.}
\label{fig2}
\end{figure*}

\section{Label-Wise Meta-Correlation Module}\label{method}

\subsection{Task Definition}
To discern the distinctions of unknown objects, inspired by GMAN \cite{zhang2022generative}, we introduce attributes embedding to our framework. Specifically, given a known or unlabeled object target with its attributes embedding, our task is to navigate to an instance of the target class with the provided attributes using visual information. For example, a known target task can be defined as ``Navigate to an AlarmClock which is `Small', `Metal', `Plastic', `Glass', `Can Be Picked up' '', while an unlabeled target task can be defined as ``Navigate to an unlabeled object which is `Big', `Ceramic', `Receptacle' ''.

Formally, let $ {C}^{k}=\left ( {c}^{k}_1, \cdots,{c}^{k}_I   \right ) $ denotes the set of $I$ known target object classes, $ {C}^{uk}=\left ( {c}^{uk}_1, \cdots,{c}^{uk}_N   \right ) $ denotes the set of $N$ unknown target object classes, and $ {C}^{us}=\left ( {c}^{us}_1, \cdots,{c}^{us}_M   \right ) $ denotes the set of $M$ unseen target object classes. These three sets have no intersection. All unknown and unseen target object classes will be designated as unlabeled objects for agent, while being distinguished based on their special attributes.

At the beginning of each episode, the agent starts from the random initial state $s=\left ( x,y,{\theta}_r,{\theta}_h \right ) $ in a random room given a target object class $to \in {C}^{k}$ or an unlabeled object target with its attributes, where $x$ and $y$ represent the position of the agent, ${\theta}_r$ and ${\theta}_h$ indicate the point of view of the agent. At each time $t$, the agent takes an action $a$ from the action set $A$ until the termination action is issued by the agent.

There are 6 types of actions in the environment, including 
\textbf{MoveAhead, RotateLeft, RotateRight, LookUp, LookDown, Done}. We define an episode as a success when the agent issues a \textbf{Done} action and (1) The target object is sufficiently close to the agent ($1.5$ meters in practice); (2) The target object is in
view; and (3) The agent does not pass the maximum number of allowed steps.

\subsection{Model Overview} 
In this section, we delve into the operational mechanics of our Zero-shot Object Navigation (ZSON) framework, the label-wise meta-correlation module (LWMCM), which is designed to enhance navigation performance by exploiting the relationships among objects with and without labels. As illustrated in Figure~\ref{fig2}, this frame work applies four key components, target feature generator (TFG, Sec.~\ref{sec:tfg}), unlabeled object identifier (UOI, Sec.~\ref{sec:uoi}), meta contrastive feature modifier (MCFM, Sec.~\ref{sec:mcfm}) and meta object-graph learner (MOGL, Sec.~\ref{sec:mogl}), to calculate the relationships among objects and produce the relationships embedding $z_r$. The relationships embedding $z_r$, concatenated with observation embedding $z_o$ and target embedding $z_t$, is then utilized to predict the action of the agent (Sec.~\ref{sec:np}).
In detail of obtaining the relationships among objects, we employ DETR to produce features of known objects $f_k$, and employ TFG, UOI and MCFM to produce the features of unlabeled objects $f_t'$, and then employ MOGL to calculate the relationships via a graph.


\begin{table*}[!htbp]
    \centering

    \caption{Details of optimizing the parameter of each component in our framework.}
    \resizebox{\linewidth}{!}{

        \begin{tabular}{|c|c|c|}
        \hline
        \textbf{Component}                       & \textbf{Pretrain} & \textbf{Training and Inference}                        \\ \hline
        \textbf{ResNet-18}                       & ImageNet          & frozen                                           \\ \hline
        \textbf{DETR, Diffusion Model}      & known objects     & frozen                                           \\ \hline
        \textbf{UOI}                             & unknown objects   & frozen                                           \\ \hline
        \textbf{MCFM}                            & None              & when target is unknown and unseen objects and UOI detected unseen target \\ \hline
        \textbf{MOGL}                            & None              & when target is known, unknown and unseen objects \\ \hline
        \textbf{$\mathbf{FFN_o}$, $\mathbf{FFN_t}$, LSTM, Done Reminder} & None              & when target is known and unknown objects         \\ \hline
        \end{tabular}
    }
    \label{table02}
\end{table*}


Specifically, the process of obtaining the features of unlabeled objects $f_t'$ consists of three step. Initially, we employ a target feature generator (TFG) to generate the features representation $g_t$ for the unlabeled target object based on its semantic attributes. However, features generated solely from semantic attributes do not account for other environmental factors like the background. To address this limitation, we introduce an unlabeled object identifier module (UOI), which takes both $g_t$ and the features $f_o$ of the observation frame as inputs. The UOI module evaluates whether an unlabeled object appeared in the frame. If so, the module output an intermediate features denoted as $f_t$ representing the unlabeled target object's features adapted to the specific environment. Then, to establish connections among object features, we employ a meta contrastive feature modifier module (MCFM). Within this module, the intermediate features $f_t$ is brought closer to the features of objects that co-occur with it in the observation frame and is pushed away from the features of objects that are not present in the frame.

As shown in Tab.~\ref{table02}, the process to update the parameters of each component can be divided into two stages. In the pretrain stage, we train component ResNet-18, DETR, Diffusion Model and UOI directly and do not conduct navigation. In the training and inference stage, we conduct navigation experiments and update the parameter MCFM, MOGL, $FFN_{o}$, $FFN_{t}$, LSTM and Done Reminder. Specifically, we train the MCFM and MOGL in a meta-learning way. The updating of parameters of these two components not only happen in training stage but also in inference stage (Sec.~\ref{sec:np}).

\begin{figure*}[!htbp]
\centering
\includegraphics[width=1\textwidth]{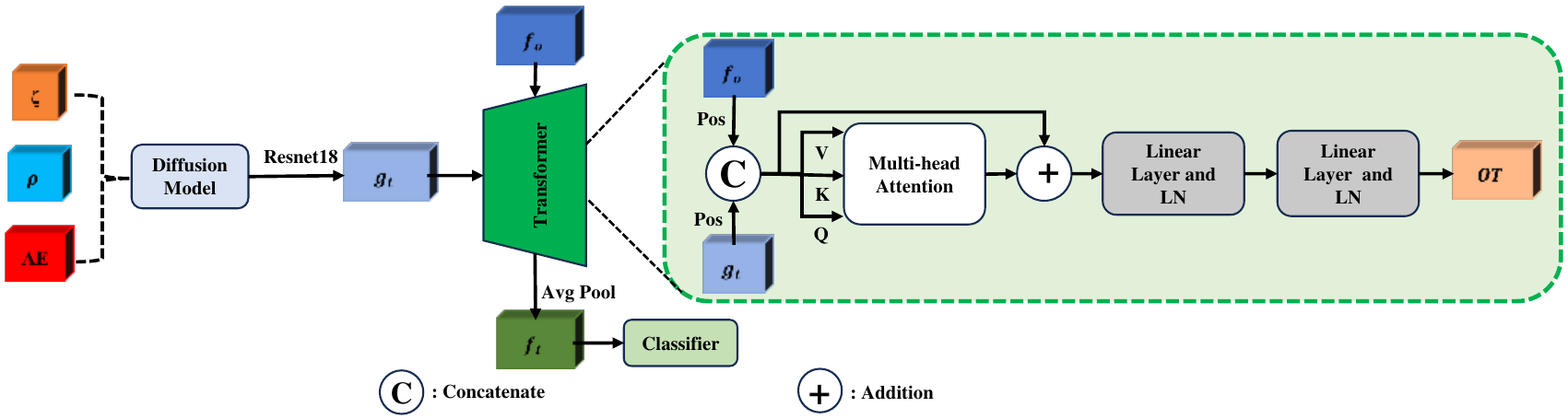}
\caption{Details of the unlabeled object identifier (UOI). The agent uses UOI to identify whether unknown or unseen objects exist in the current field of view based on the target generative features $g_t$ and the observation features $f_o$.}
\label{fig3}
\end{figure*}

\subsection{Target Feature Generator} \label{sec:tfg}
Mastering the projection from semantic attributes to visual features, thereby equipping the agent with the capability to gather insights about unlabeled objects, including unknown and unseen objects, is pivotal for successful navigation. Drawing inspiration from GMAN \cite{zhang2022generative}, we utilize a target feature generator (TFG) pre-trained to adeptly generate visual representations for target classes. The TFG consists of a generator G that generate images for unknown or unseen objects. Starting from a sampled noisy image and a provided object attribute description, the diffusion model performs $\zeta $ denoising steps until a clear image is formed. Let $\hbar  \in {C}^{k}\cup {C}^{uk}\cup {C}^{us}$ denotes the target category, $\rho $ denotes a sampled noisy image from a Gaussian prior, and $AE$ denotes the attributes embedding function.
The generative procedure can be regarded as $G\left (  \left ( \rho , AE({\hbar }) \right ), \zeta  \right ) $. During the training phase, we use attributes embedding of known objects and the corresponding pre-collected images to learn the projection between semantic and visual information. In the generation phase, we use attributes combinations of unknown or unseen objects to generate corresponding images. This pretraining process is similar with \cite{liu2022compositional}. In particular, for generative images ${g_i}=G\left (  \left ( \rho , AE({\ell}) \right ), \zeta  \right )$ of unknown or unseen object classes $\ell  \in {C}^{uk}\cup {C}^{us}$, we extract them by ResNet18 pre-trained on ImageNet to obtain the unknown or unseen object generative features ${g_t}\in {R}^{{d}_g\times 512}$. ${d}_g$ is the pixel number of the image feature map.

\subsection{Unlabeled Object Identifier}\label{sec:uoi}

To facilitate the agent's perception of the existence of unknown or unseen objects within its current visual field, we pretrain an unlabeled object identifier module. Specially, let ${f_o} \in {R}^{{d}_g\times 512}$ (from ResNet18) denotes the observation features of each view and ${g_t}^{uk}\in {R}^{N \times {d}_g\times 512}$ denotes the generative features of N unknown object classes. Note that we only use unknown object classes for UOI pretraining. Motivated by \cite{tan2021instance}, transformer is employed to learning between current features and generative features. As shown in Figure~\ref{fig3}, we define the input as :
\begin{equation}
\tilde{f_o}=Pos(f_o)
\end{equation}

\begin{equation}
{\tilde{g_t}}^{uk}[n, :]=Pos({g_t}^{uk}[n, :])
\end{equation}

\begin{equation}
{X}^{n}=Concat(\tilde{f_o}, {\tilde{g_t}}^{uk}[n, :])
\end{equation}
Here, $n=0,\cdot \cdot \cdot ,N-1$ is the index of N unknown object classes. $Pos$ is the position embedding \cite{carion2020end} function. With input ${X}^{n} \in {R}^{2{d}_g\times 512}$, we employ a total of Y transformer layers. For the $n$-th unknown object class, we denote ${OT}^{n} \in {R}^{512}$ which is extracted from the last transformer layer as the output of transformer. 
Then, the final classifier is formulated as:
\begin{equation}
f_t = \delta((\frac{1}{N} \sum_{i=1}^{N}{{OT}^{i}}) {W}^{{M}_1})
\end{equation}

\begin{equation}
CLS = f_t{W}^{{M}_2}
\end{equation}
where $\delta$ denotes the ReLU function, $\frac{1}{N} \sum_{i=1}^{N}{{OT}^{i}}$ is squeezing all unknown object classes information into a descriptor using global average pooling. ${W}^{{M}_1} \in {R}^{512\times 256}$ and ${W}^{{M}_2} \in {R}^{256\times 1}$ are two linear functions. $CLS$ is the final output which equals to one when unknown objects present in current field of view, or zero otherwise. Let $gt$ denotes the ground truth label of current view. Finally, we pretrain our model to optimize a cross entropy loss:

\begin{equation}
L_{uoi}=CE(CLS, gt)
\end{equation}

\begin{figure*}[!htbp]
\centering
\includegraphics[width=1\textwidth]{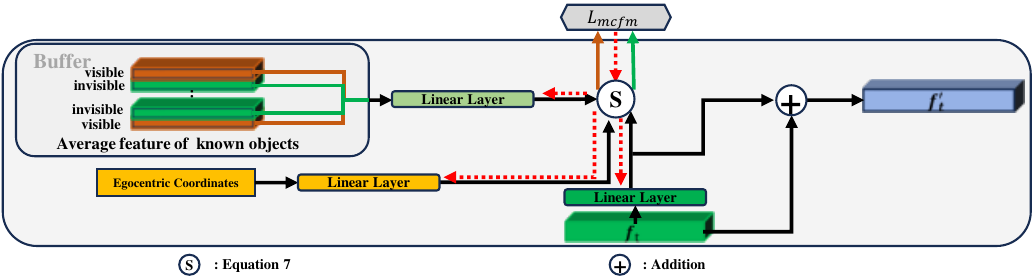}
\caption{Details of the meta contrastive feature modifier (MCFM). The immediate features $f_t$ was brought closer to the
features of objects that co-occur with it in the observation
frame and was pushed away from the features of objects that
are not present in the frame using function $S$ and $l_{mcfm}$.
}
\label{fig4}
\end{figure*}

\subsection{Meta Contrastive Feature Modifier}\label{sec:mcfm}

In the above section, we mention that the learnable features vector $f_t \in {R}^{1\times 256}$ obtained through global average pooling and a linear function in UOI serves as an unified features representation of unknown or unseen object classes. This informative vector can enhance the capacity of searching and navigating. Besides, we have empirically observed that the occurrence of diverse unknown objects is contingent upon specific observation positions and context. For example, an unknown object phone often appears simultaneously with a known object laptop on the sofa or table. In order to establish connections between known and unlabeled objects, we draw inspiration from this and propose a novel meta contrastive feature modifier module (MCFM) that integrates positional and contextual factors as essential constraints for learning. Figure~\ref{fig4} shows details of our MCFM module. Specially in the current view, let $CLS$ denotes the output of UOI , $f_k^c \in {R}^{1\times 256}$ denotes the features of category $c$ extracted from a detector DETR \cite{carion2020end} and $c \in {C}^{k}$, $p \in {R}^{1\times 6}$ denotes the egocentric coordinates through Egocentric Coordinate Transformation by DAT \cite{dang2022search}, and $SV(c)$ is a function which equals to one when both the known object $c$ and unknown objects appear in the current view, or zero otherwise. Note that we treat $f_k^c$ as contextual factors and employ a buffer to store $f_k^c$ by calculating the mean of all observed features of category $c$. Formally, we design the following function $S$:

\begin{equation}
S(f_k^c,p,f_t)=\left \|f_k^c{W}^{{R}_1} + p{W}^{{R}_2} - f_t{W}^{{R}_3}\right \|_{2}^{2}
\end{equation}


which is designed inspired by \cite{bordes2013translating}, and ${W}^{{R}_1} \in {R}^{256\times 256}$, ${W}^{{R}_2} \in {R}^{6\times 256}$ and ${W}^{{R}_3} \in {R}^{256\times 256}$ are learnable parameters. Our objective is to bring $f_t$ closer to features of objects that co-occur with it and push it away from features of objects that are not present in the frame. Based on this, let $O=\left \{ c|SV(c)=1,CLS=1,c \in {C}^{k} \right \}$, and $\hat{O}=\left \{ \hat{c}|SV(\hat{c})=0,CLS=1,\hat{c} \in {C}^{k} \right \}$, and we set the following loss:

\begin{equation}
\begin{split}
{L}_{mcfm}=-\ln\vartheta(\frac{1}{|\hat{O}|}\sum_{\hat{c} \in \hat{O}}S(f_k^{\hat{c}},p,f_t) - 
\\
\frac{1}{|O|}\sum_{c \in O}S(f_k^c,p,f_t))
\end{split}
\end{equation}

Here $\vartheta(\cdot )$is the sigmoid function. Finally, we define the output of MCFM as:
\begin{equation}
f_t'=f_t + f_t{W}^{{R}_3}
\label{ft}
\end{equation}

\begin{figure*}[t]
\centering
\includegraphics[width=1\linewidth]{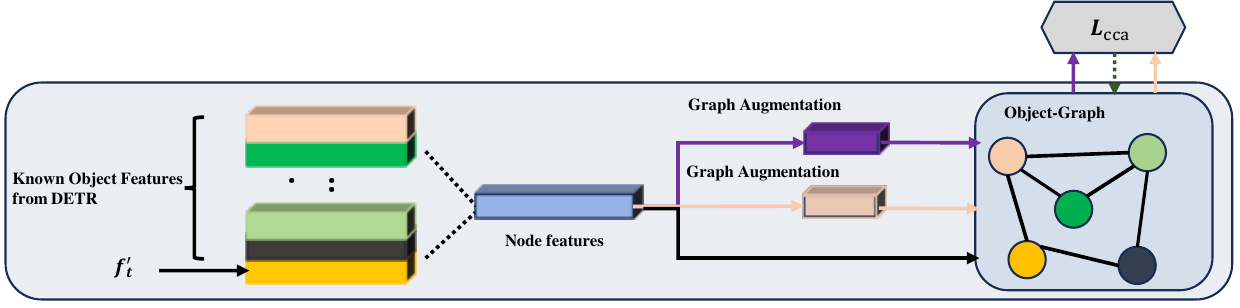}
\caption{Details of the meta object-graph learner (MOGL). The joint features of known object features from DETR and modified $f_t'$ from MCFM perform as node features of the Object-Graph. Then the CCA-SSG method is used to capture more informative features among these node features.}
\label{fig5}
\end{figure*}

\subsection{Meta Object-Graph Learner}\label{sec:mogl}

To efficiently capture informative object features and enable more nuanced comprehension of their contextual relationships, we employ a meta object-graph learner (MOGL) with a self-supervised method. We first denote a graph $G=(V,E)$, where $V$ and $E$ denote the nodes and the edges between nodes respectively. As shown in Figure~\ref{fig5}, each node $v \in V$ denotes features representation with known objects obtained through DETR and unlabeled objects derived from the output $f_t'$ of MCFM, and each edge $e \in E$ denotes the proximity relationships between objects. Then graph convolutional network (GCN) takes all nodes as inputs $V \in {R}^{(I+1)\times 256}$ and embeds each node for the following LSTM.

\begin{equation}
F=\delta(E \cdot V \cdot {W}^{G})
\end{equation}

where $F$ is the encoded $G$ and ${W}^{G}$ is the learnable parameters of GCN. Moreover, inspired by \cite{zhang2021canonical}, we leverage the CCA(canonical correlation analysis)-SSG method to enhance node representation and capture more informative features for the following LSTM. Formally, let ${F}_{A}$ and ${F}_{B}$ denote the embedded graphs of two views after random graph augmentations of graph $G$, $\eta $ denotes a non-negative hyperparameter and $U$ represents the identity matrix. According to CCA-SSG, we set the following self-supervised loss:

\begin{equation}
\begin{split}
{L}_{cca}=\left \| {F}_{A}-{F}_{B} \right \|_{2}^{2} + \eta (\left \| {F}_{A}^{T}{F}_{A}-U \right \|_{2}^{2} 
\\
+ \left \| {F}_{B}^{T}{F}_{B}-U \right \|_{2}^{2})
\end{split}
\label{cca}
\end{equation}

\begin{algorithm}[tb]
    \caption{Training stage}
    \label{alg:algorithm}
    \textbf{Input}:  Randomly initial parameters $\alpha$ for MCFM, $\beta$ for MOGL and $\psi$ for the remaining components of the model.The learning rate ${\lambda}_{1}$, ${\lambda}_{2}$, $\mu$. The training tasks ${\Gamma }_{train}$.
    \begin{algorithmic}[1]
    \WHILE{not converged}
    \FOR{mini-batch of tasks ${\tau}_{i} \in {\Gamma }_{train}$}
    \STATE ${\alpha}_{i} \leftarrow \alpha$
    \STATE ${\beta}_{i} \leftarrow \beta$
    \WHILE{termination action is not issued}
    \STATE Obtain $CLS$ from current view
    \STATE Take action $a$ sampled from ${\pi}_{{\alpha}_{i}, {\beta}_{i}, {\psi}}$
    \IF {$CLS=1$}
    \STATE ${\alpha}_{i} \leftarrow {\alpha}_{i} - {\lambda}_{1}{\bigtriangledown}_{{\alpha}_{i}}{L}_{mcfm}({\alpha}_{i})$
    \ENDIF
    \STATE ${\beta}_{i} \leftarrow {\beta}_{i} - {\lambda}_{2}{\bigtriangledown}_{{\beta}_{i}}{L}_{cca}({\alpha}_{i}, {\beta}_{i},\psi)$
    \ENDWHILE
    \ENDFOR
    \STATE $\beta \leftarrow \beta - {\mu} {\textstyle \sum_{i}} {\bigtriangledown}_{\beta}{L}_{a3c}({\alpha}_{i}, {\beta}_{i},\psi)$
    \STATE $\psi \leftarrow \psi - {\mu} {\textstyle \sum_{i}} {\bigtriangledown}_{\psi}{L}_{a3c}({\alpha}_{i}, {\beta}_{i},\psi)$
    \ENDWHILE
    \end{algorithmic}
    \textbf{Output} $\alpha, \beta, \psi$
\end{algorithm}

\subsection{Navigation Policy}\label{sec:np}
    Our target indicator $TI$ is concatenated by a one-hot encoded target vector (including $I$ known object classes and an unknown label) and attributes embedding of current target. After obtaining the visual features $f_o$, the output $F$ of MOGL and the target indicator $TI$, we project them into a space with a same dimension using Feed Forward Network (FFN) and concatenate them with previous actions and states. Then we take them as the input of the Long Short-Term Memory (LSTM) \cite{hochreiter1997long} and get ${S}_{t}$. Following previous works \cite{du2020learning, zhang2022generative}, we take ${S}_{t}$ as the input of A3C architecture \cite{mnih2016asynchronous} to train the policy network $\pi({a}_{t}|{S}_{t}, TI)$. Motivated by \cite{zhang2021hierarchical}, a done reminder is also employed to the network to remind agent to stop in time when it encounters the known objects or unknown label of target. At each time, the agent selects an action with the highest probability  and uses the predicted value to train the policy network.
    
    Overall, our total loss is set as equation \ref{totalloss}. Algorithm~\ref{alg:algorithm} shows the details of the training phase. Specially, MCFM (with parameter $\alpha$) is updated with $L_{mcfm}$ only when unknown or unseen objects appear in view during navigation. MOGL (with parameter $\beta$) is updated with $L_{cca}$ and $L_{a3c}$ and the remaining components of the model (with parameter $\psi$) is updated with $L_{a3c}$ over all training episodes.

    Besides, we introduce the meta-learning method into our framework. It provides an algorithm to acquire the necessary adaptation skills and effectively generalize to fit the sub-tasks within a small number of adaptation steps. In our work, we regard each episode in navigation as a new task. Specially in Algorithm~\ref{alg:algorithm}, the line 9 and 11 is the inner-loop executed in both training and inference, and the line 14 and 15 is the outer-loop that only conducted in training. 
    \begin{equation}
    L={\lambda}_{1}{L_{mcfm}} + {\lambda}_{2}{L_{cca}} + \mu{L_{a3c}}
    \label{totalloss}
    \end{equation}

\begin{table*}[t]
\centering
\caption{ Comparison with the related methods (\%) in unseen environments on AI2THOR simulator.}
\resizebox{\linewidth}{!}{
\begin{tabular}{c|cccccc}
\hline
\multirow{2}{*}{Methods} & \multicolumn{2}{c|}{Known Objects} & \multicolumn{2}{c|}{Unknown Objects} & \multicolumn{2}{c}{Unseen Objects} \\ \cline{2-7}
 & SR & \multicolumn{1}{c|}{SPL} & SR & \multicolumn{1}{c|}{SPL} & SR & SPL \\
\hline
\hline
Random & $4.53_{\pm 0.84}$ & \multicolumn{1}{c|}{$2.02 _{\pm 0.31}$} & $0.80 _{\pm 0.07}$ & \multicolumn{1}{c|}{$0.22 _{\pm 0.31}$} & $0.00 _{\pm 0.00}$ & $0.00 _{\pm 0.00}$\\

Baseline & $\mathbf{52.20_{\pm 1.10}}$ & \multicolumn{1}{c|}{$\mathbf{33.40_{\pm 0.58}}$} & $20.80_{\pm 0.66}$ & \multicolumn{1}{c|}{$9.28_{\pm0.84}$} & $18.50_{\pm0.13}$ & $8.05_{\pm0.29}$ \\

SP \cite{yang2018visual} & $45.30_{\pm 0.78}$ & \multicolumn{1}{c|}{$24.80_{\pm1.22}$} & $20.50_{\pm0.82}$ & \multicolumn{1}{c|}{$10.11_{\pm0.21}$} & $19.60_{\pm0.08}$ & $9.86_{\pm0.10}$ \\
SAVN \cite{wortsman2019learning} & $48.60_{\pm 1.29}$ & \multicolumn{1}{c|}{$27.20_{\pm1.15}$} & $19.20_{\pm0.75}$ & \multicolumn{1}{c|}{$6.21_{\pm0.19}$} & $18.60_{\pm0.10}$ & $5.96_{\pm0.25}$ \\

\hline
Ours & $50.70_{\pm1.20}$ & \multicolumn{1}{c|}{$32.65_{\pm0.48}$} & $\mathbf{40.40_{\pm1.52}}$ & \multicolumn{1}{c|}{$\mathbf{29.47_{\pm0.47}}$} & $\mathbf{38.50_{\pm1.44}}$ & $\mathbf{27.02_{\pm1.25}}$ \\
\hline
\end{tabular}%
}

\label{table2}
\end{table*}

\begin{table*}[t]
\centering
\caption{ Comparison with the related methods (\%) in unseen environments on RoboTHOR simulator.}
\resizebox{\linewidth}{!}{
\begin{tabular}{c|cccccc}
\hline
\multirow{2}{*}{Methods} & \multicolumn{2}{c|}{Known Objects} & \multicolumn{2}{c|}{Unknown Objects} & \multicolumn{2}{c}{Unseen Objects} \\ \cline{2-7}
 & SR & \multicolumn{1}{c|}{SPL} & SR & \multicolumn{1}{c|}{SPL} & SR & SPL \\
\hline
\hline
Random & $2.59_{\pm 0.44}$ & \multicolumn{1}{c|}{$1.02 _{\pm 0.11}$} & $0.00 _{\pm 0.00}$ & \multicolumn{1}{c|}{$0.00 _{\pm 0.00}$} & $0.00 _{\pm 0.00}$ & $0.00 _{\pm 0.00}$\\

Baseline & $\mathbf{42.50_{\pm 0.27}}$ & \multicolumn{1}{c|}{$\mathbf{23.87_{\pm 0.50}}$} & $17.60_{\pm 0.79}$ & \multicolumn{1}{c|}{$9.52_{\pm0.14}$} & $16.70_{\pm0.58}$ & $9.15_{\pm0.40}$ \\

SP \cite{yang2018visual} & $20.20_{\pm 1.21}$ & \multicolumn{1}{c|}{$12.55_{\pm0.47}$} & $15.70_{\pm0.20}$ & \multicolumn{1}{c|}{$10.01_{\pm0.41}$} & $13.20_{\pm0.85}$ & $7.72_{\pm0.50}$ \\
SAVN \cite{wortsman2019learning} & $32.90_{\pm 0.50}$ & \multicolumn{1}{c|}{$16.80_{\pm1.15}$} & $18.50_{\pm0.21}$ & \multicolumn{1}{c|}{$10.21_{\pm0.88}$} & $20.90_{\pm0.33}$ & $11.26_{\pm0.74}$ \\
GMAN \cite{zhang2022generative} & $37.10_{\pm 0.61}$ & \multicolumn{1}{c|}{$16.80_{\pm1.15}$} & - & \multicolumn{1}{c|}{-} & $27.67_{\pm0.67}$ & $14.29_{\pm0.37}$ \\

\hline
Ours & $40.20_{\pm1.31}$ & \multicolumn{1}{c|}{$21.22_{\pm1.08}$} & $\mathbf{31.20_{\pm1.55}}$ & \multicolumn{1}{c|}{$\mathbf{20.10_{\pm0.11}}$} & $\mathbf{29.50_{\pm0.85}}$ & $\mathbf{19.82_{\pm0.97}}$ \\
\hline
\end{tabular}%
}

\label{table3}
\end{table*}

\section{Experiments} \label{experiment}
\subsection{Experimental Setup}
We choose AI2THOR \cite{kolve2017ai2} and RoboTHOR \cite{deitke2020robothor} environment to evaluate our method. 
We first modify the two mentioned simulators to ensure the presence of only known and unknown objects in the training scenes, while including known, unknown and unseen objects in the validation and testing scenes. AI2THOR includes 30 different rooms and we use 20 rooms for training, 5 rooms for validation, and 5 rooms for testing. RoboTHOR consists of 75 apartments for training and validation and we choose 60 apartments for training, 5 for validation and 10 for testing. The details of the split of target objects are presented in Table~\ref{suptable1}.

\subsection{Implementation Details}
We train our model 6M episodes with 18 asynchronous workers on 2 RTX 3090 Nvidia GPUs. The number of transformer layers in UOI is 8. The hyperparameter in equation~\ref{cca} is set to ${10}^{-3}$ according to \cite{zhang2021canonical}. The learning rate $({\lambda}_{1}$, ${\lambda}_{2}$, $\mu$) in Algorithm~\ref{alg:algorithm} is all set to ${10}^{-4}$.

\subsection{Evaluation Metrics}
We use the success rate (SR), success weighted by path length (SPL) \cite{anderson2018evaluation} to evaluate our method. SR refers to
the success rate of agent in finding the target object and is formulated
as $SR=\frac{1}{F} {\textstyle \sum_{i=1}^{F}} {Suc}_{i}$ , where $F$ is the number of episodes and ${Suc}_{i}$ indicates whether the $i$-th episode succeeds. SPL considers both the success rate and the path length and is defined as $SPL=\frac{1}{F}{\textstyle \sum_{i=1}^{F}}{Suc}_{i}\frac{{L}_{i}^{*}}{max({L}_{i}, {L}_{i}^{*})}$, where ${L}_{i}$
is the path length taken by the agent and ${L}_{i}^{*}$ is the theoretical shortest path.

\subsection{Comparison Methods}
We compare the following methods:

\textbf{Baseline}. We feed the current egocentric RGB image embedding, target indicator and the detection results of DETR to LSTM and use A3C architecture for navigation as our baseline. 

\textbf{Random policy}. The agent will randomly walk and stop in the scene.

\textbf{SP}. SP \cite{yang2018visual} constructs additional scene prior knowledge encoded with GCN from an external dataset to assist the agent.

\textbf{SAVN}. SAVN \cite{wortsman2019learning} incorporates a sub-network to learn a self-supervised interaction loss, utilizes meta reinforcement learning for the agent to adapt to unseen environments during inference, and employs GloVe embedding \cite{pennington2014glove} to associate target appearance and concept.

\textbf{GMAN}. GMAN \cite{zhang2022generative} use generative adversarial network \cite{goodfellow2014generative} to generate visual features of unseen objects based on their category semantic embedding and fit the environment adaptively within the meta-learning structure.

\subsection{Results}
In our evaluation conducted on the AI2THOR and RoboTHOR test datasets, as detailed in Table~\ref{table2} and Table~\ref{table3}, we performed experiments in triplicate, providing mean values and standard deviations for a robust statistical analysis. Our methodology strategically omits unknown objects from the baseline and related work comparisons to underscore the significance of including unknown objects for enhancing navigation towards unseen targets. Our findings reveal that our approach significantly surpasses existing methods, achieving improvements of 19.6\% in Success Rate (SR) and 19.36\% in Success weighted by Path Length (SPL) in AI2THOR, and 12.7\% in SR and 9.89\% in SPL in RoboTHOR for unknown object navigation. Similarly, for unseen object navigation, our method registers substantial gains of 18.9\% in SR and 17.16\% in SPL in AI2THOR, and 1.83\% in SR and 5.53\% in SPL in RoboTHOR. However, we observed a negative transfer effect wherein the integration of unknown object data during training slightly impedes the SR for known object navigation, suggesting that learning from unknown objects might inadvertently influence the agent's performance negatively on familiar tasks. This phenomenon underscores the complex dynamics between learning from unknown versus known objects.

The variability in training and testing object characteristics—such as size, visibility, and spatial distribution—plays a crucial role in shaping the experimental results. Given the unique target selection in AI2THOR and the scalability challenges with GMAN, we abstain from direct comparisons with GMAN in the AI2THOR environment. However, by aligning our target selection with GMAN in the RoboTHOR environment, we ensure a more reliable and equitable comparison, where our method demonstrates commendable performance.

\subsection{Case Study}
\begin{figure}[t]
\centering
\includegraphics[width=\linewidth]{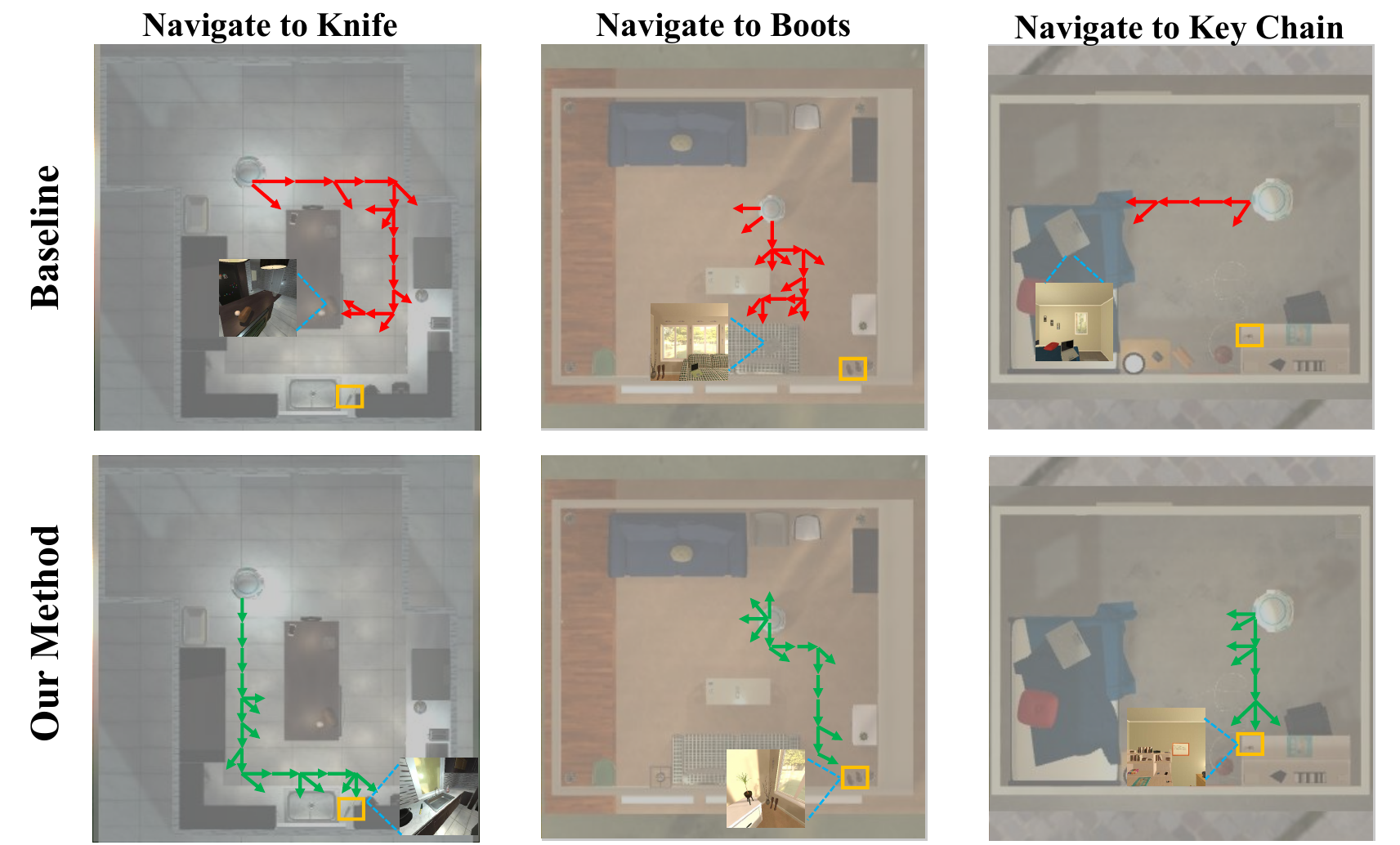}
\caption{Visualization of unseen object navigation in testing environments. The trajectory of the agent is indicated by red and green arrows and the target objects are highlighted by the orange boxes.}
\label{fig7}
\end{figure}

Figure~\ref{fig7} illustrates the navigation paths from three separate episodes, comparing the baseline method to our proposed approach. In the baseline scenarios, the agent's decision-making is heavily skewed towards known objects, as seen when the agent, tasked with locating an unknown object like a knife, mistakenly concludes its mission upon encountering a known object such as bread. This pattern is consistent across the other examples provided. Conversely, our method, depicted in the lower row of the figure, significantly enhances the agent's ability to identify and navigate towards unknown objects, demonstrating a marked improvement in navigation efficiency and accuracy.

\begin{table*}[t]
\centering
\caption{Ablation experiments (\%) for each components on AI2THOR simulator. UOT represents ``Unknown Object Targets''(i.e. employing unknown objects as training targets).}
\resizebox{\linewidth}{!}{%
\begin{tabular}{ccccc|cccccc}
\hline
\multirow{2}{*}{UOT} & \multirow{2}{*}{TFG} & \multirow{2}{*}{UOI} & \multirow{2}{*}{MCFM} & \multirow{2}{*}{MOGL} & \multicolumn{2}{c|}{Known Objects} & \multicolumn{2}{c|}{Unknown Objects} & \multicolumn{2}{c}{Unseen Objects} \\ \cline{6-11}
 &  &  &  &  & SR & \multicolumn{1}{c|}{SPL} & SR & \multicolumn{1}{c|}{SPL} & SR & SPL \\ \hline \hline
 
 &  &  &  &  & $\mathbf{52.20_{\pm 1.10}}$ & \multicolumn{1}{c|}{$\mathbf{33.40_{\pm 0.58}}$} &  $20.80_{\pm 0.66}$ & \multicolumn{1}{c|}{$9.28_{\pm 0.84}$} & $18.50_{\pm 0.13}$ & $8.05_{\pm 0.29}$ \\
 
 \checkmark &  &  &  &  & $49.40_{\pm 0.18}$ & \multicolumn{1}{c|}{$29.29_{\pm 0.12}$} &  $29.30_{\pm 0.05}$ & \multicolumn{1}{c|}{$23.03_{\pm 0.24}$} & $28.00_{\pm 0.85}$ & $21.88_{\pm 0.14}$ \\

 \checkmark &  \checkmark &  \checkmark &  &  & $48.40_{\pm 0.30}$ & \multicolumn{1}{c|}{$28.46_{\pm 0.45}$} &  $33.00_{\pm 0.09}$ & \multicolumn{1}{c|}{$24.67_{\pm 0.41}$} & $31.40_{\pm 0.10}$ & $22.07_{\pm 0.32}$ \\
 
 \checkmark &  \checkmark &  \checkmark &  \checkmark &  & $47.50_{\pm 0.27}$ & \multicolumn{1}{c|}{$28.20_{\pm 0.88}$} & $35.50_{\pm 0.40}$ & \multicolumn{1}{c|}{$27.32_{\pm 0.28}$} & $32.20_{\pm 0.44}$ & $25.01_{\pm 0.30}$ \\ \hline
 
 \checkmark &  \checkmark &  \checkmark &  \checkmark &  \checkmark & $50.70_{\pm 1.20}$ & \multicolumn{1}{c|}{$32.65_{\pm 0.48}$} &  $\mathbf{40.40_{\pm 1.52}}$ & \multicolumn{1}{c|}{$\mathbf{29.47_{\pm 0.47}}$} & $\mathbf{38.50_{\pm 1.44}}$ & $\mathbf{27.02_{\pm 1.25}}$ \\ \hline
\end{tabular}%
}

\label{table4}
\end{table*}

\begin{table*}[t]
\centering
\caption{Ablation experiments (\%) for loss function and meta-learning setting of MCFM and MOGL on AI2THOR simulator.}
\resizebox{\linewidth}{!}{
\begin{tabular}{c|cc|cc|cc}
\hline
\multirow{2}{*}{Method} & \multicolumn{2}{c|}{Known Objects} & \multicolumn{2}{c|}{Unknown Objects} & \multicolumn{2}{c}{Unseen Objects} \\ \cline{2-7} 
 & SR & SPL & SR & SPL & SR & SPL \\ \hline \hline
 MCFM $\xrightarrow{}$ No $L_{mcfm}$& $50.20_{\pm 1.08}$ & $31.62_{\pm 0.33}$ & $37.10_{\pm 0.27}$ & $27.07_{\pm 0.70}$ & $35.30_{\pm 0.42}$ & $25.98_{\pm 1.01}$ \\
 MOGL $\xrightarrow{}$ No $L_{cca}$& $48.90_{\pm 0.88}$ & $30.40_{\pm 0.45}$ & $39.50_{\pm 0.89}$ & $26.55_{\pm 1.25}$ & $37.80_{\pm 0.23}$ & $25.27_{\pm 1.31}$ \\
 MCFM $\xrightarrow{}$ No meta& $50.60_{\pm 1.10}$ & $32.35_{\pm 0.55}$ & $37.50_{\pm 0.43}$ & $28.60_{\pm 0.22}$ & $36.40_{\pm 0.42}$ & $24.73_{\pm 0.08}$ \\
 MOGL $\xrightarrow{}$ No meta& $49.70_{\pm 0.25}$ & $32.07_{\pm 0.88}$ & $40.00_{\pm 1.22}$ & $28.97_{\pm 0.59}$ & $38.30_{\pm 0.91}$ & $25.84_{\pm 1.20}$ \\ \hline
\multicolumn{1}{l|}{Complete Method} & \multicolumn{1}{l}{$\mathbf{50.70_{\pm 1.20}}$} & \multicolumn{1}{l|}{$\mathbf{32.65_{\pm 0.48}}$} & \multicolumn{1}{l}{$\mathbf{40.40_{\pm 1.52}}$} & \multicolumn{1}{l|}{$\mathbf{29.47_{\pm 0.47}}$} & \multicolumn{1}{l}{$\mathbf{38.50_{\pm 1.44}}$} & \multicolumn{1}{l}{$\mathbf{27.02_{\pm 1.25}}$} \\ \hline
\end{tabular}%
}

\label{table5}
\end{table*}

\subsection{Ablation Study}

\subsubsection{Impacts of Different Components}
Our ablation study, detailed in Table~\ref{table4}, scrutinizes the impact of each component within our methodology, revealing that the integration of unknown object targets (UOT) as an additional training dimension notably boosts baseline capabilities in navigating towards both unknown and unseen objects. This enhancement underlines the augmented proficiency of the agent in addressing novel objects. Furthermore, the meta contrastive feature modifier (MCFM) plays an indispensable role by refining the feature representations of unknown objects, synthesizing inputs from both the target feature generator (TFG) and the unlabeled object identifier (UOI). The inclusion of MCFM's output during training markedly elevates performance in navigating towards unknown and unseen objects. The introduction of the meta object-graph learner (MOGL) to formulate an object relationship graph further empowers the agent with the ability to navigate more adeptly within intricate settings. Collectively, these components significantly elevate our method's efficiency in unknown and unseen object navigation, achieving substantial enhancements over the baseline approach.

\subsubsection{Impacts of Loss Function and Meta-Learning Setting}
Table~\ref{table5}  outlines the results of ablation experiments focused on the loss function and meta-learning configuration for the meta contrastive feature modifier (MCFM) and meta object-graph learner (MOGL). The enhancements observed are initially driven by $L_{mcfm}$, which forges vital links between seen and unseen objects, enriching the navigation model's contextual understanding. Furthermore, the implementation of the CCA-SSG technique within MOGL significantly boosts node representation, capturing more detailed and informative features essential for nuanced navigation. Additionally, our framework's meta-learning approach plays a pivotal role, enabling dynamic adjustments to visual representations during the test phase. This flexibility is key to the agent's success in navigating towards unseen objects within novel environments, thereby markedly improving its overall performance.

\begin{table*}[t]
\centering
\caption{Distribution of the target objects in AI2THOR and RoboTHOR.}
\resizebox{1\textwidth}{!}{%
\begin{tabular}{c|llll|ll|ll}
\hline
\multicolumn{1}{l|}{Simulator} & \multicolumn{4}{c|}{Known Objects} & \multicolumn{2}{c|}{Unknown Objects} & \multicolumn{2}{c}{Unseen Objects} \\ \hline
\multirow{5}{*}{AI2THOR} & AlarmClock & BasketBall & Bowl & Bread & Bathtub & Book & Apple & BaseballBat \\
 & Chair & CreditCard & DeskLamp & Faucet & FloorLamp & Microwave & KeyChain & Knife \\
 & Fridge & GarbageCan & HousePlant & LightSwitch & Television & Toaster & Boots & Cup \\
 & Kettle & Laptop & Pillow & RemoteControl & CellPhone & CoffeeMachine & Pot & SoapBar \\
 & SoapBottle & Sink & Toilet & ToiletPaper & Sofa & TeddyBear \\ \hline
\multicolumn{1}{l|}{\multirow{3}{*}{RoboTHOR}} & Book & Bowl & Chair & Mug & CellPhone & DeskLamp & ArmChair & Plate \\
\multicolumn{1}{l|}{} & SideTable & Laptop & Box & Shelf & Newspaper & & Pot & Drawer \\
\multicolumn{1}{l|}{} & Television & Sofa &  &  &  &  &  &  \\ \hline
\end{tabular}%
}

\label{suptable1}
\end{table*}

\begin{table*}[t]
\centering
\caption{ Comparison with the corresponding ground truth (GT) of the the UOI output $CLS$ in unseen environments on AI2THOR simulator (\%).}
\resizebox{1\textwidth}{!}{
\begin{tabular}{c|cccccc}
\hline
\multirow{2}{*}{Methods} & \multicolumn{2}{c|}{Known Objects} & \multicolumn{2}{c|}{Unknown Objects} & \multicolumn{2}{c}{Unseen Objects} \\ \cline{2-7}
 & SR & \multicolumn{1}{c|}{SPL} & SR & \multicolumn{1}{c|}{SPL} & SR & SPL \\
\hline
\hline
Ours & $50.70_{\pm1.20}$ & \multicolumn{1}{c|}{$32.65_{\pm0.48}$} & $40.40_{\pm1.52}$ & \multicolumn{1}{c|}{$29.47_{\pm0.47}$} & $38.50_{\pm1.44}$ & $27.02_{\pm1.25}$ \\
Ours(GT) & $50.58_{\pm0.99}$ & \multicolumn{1}{c|}{$32.01_{\pm0.67}$} & $\mathbf{44.08_{\pm0.58}}$ & \multicolumn{1}{c|}{$\mathbf{30.85_{\pm0.22}}$} & $\mathbf{42.35_{\pm0.66}}$ & $\mathbf{27.09_{\pm0.85}}$ \\
\hline
\end{tabular}%
}

\label{suptable2}
\end{table*}

\begin{table}[t]
\centering
\caption{Experiments to analyse distance to targets in unseen environments on AI2THOR simulator (\%). $L\ge 5$ indicates the shortest distance is larger than 5 steps.}
\resizebox{1\columnwidth}{!}{
\begin{tabular}{c|cccc}
\hline
\multirow{2}{*}{Methods} & \multicolumn{2}{c|}{$L\ge 1$} & \multicolumn{2}{c}{$L\ge 5$}
\\ 
\cline{2-5} & SR & \multicolumn{1}{c|}{SPL} & SR & SPL \\
\hline
\hline
Unknown Targets & $40.40_{\pm1.52}$ & \multicolumn{1}{c|}{$29.47_{\pm0.47}$} & $14.50_{\pm0.33}$ & $9.52_{\pm0.11}$ \\
Unseen Targets & $38.50_{\pm1.44}$ & \multicolumn{1}{c|}{$27.02_{\pm1.25}$} & $11.84_{\pm0.22}$ & $8.37_{\pm0.05}$ \\
\hline
\end{tabular}%
}
\label{suptable3}
\end{table}

\begin{figure}[t]
\centering
\includegraphics[width=1\columnwidth]{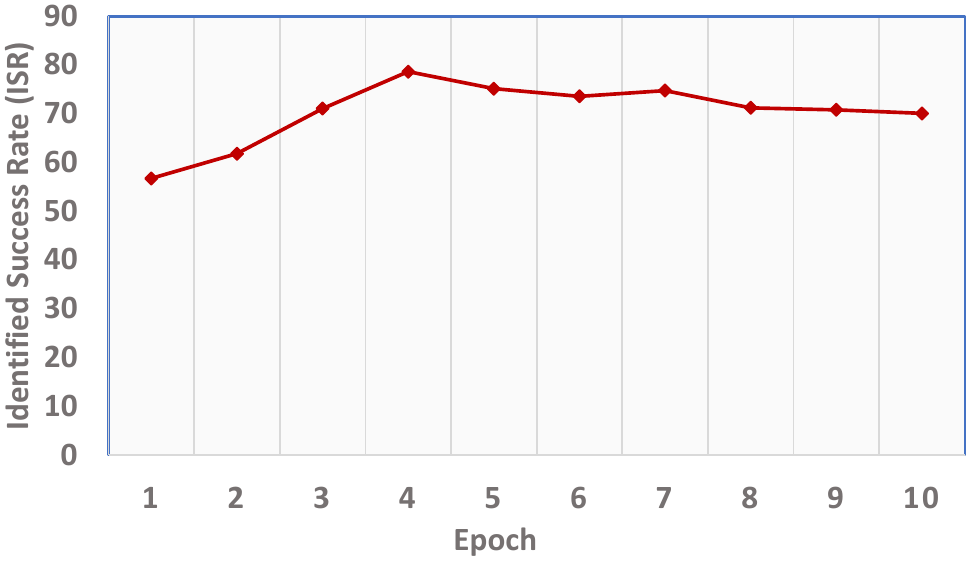}
\caption{Identified Success Rate (ISR) of unlabeled object identifier (UOI) Pre-training. }
\label{sup1}
\end{figure}

\subsection{Additional Study}
\subsubsection{Impact of UOI Pre-training on Success Rate}
In our experiment, we trained UOI with 10 epochs in training scenes of modified simulations, and we conducted tests in testing scenes to calculate the Identified Success Rate (ISR). As shown in Figure \ref{sup1}, we choose the UOI of the forth epoch exhibiting the highest ISR and integrate it into our navigation network. Higher ISR means that the agent has better discernment ability for unknown or unseen objects. Nevertheless, to assess the impact of UOI on the upper performance limit of our approach, we substituted the UOI output $CLS$ with the corresponding ground truth and employed it in our navigation network. The result is shown in Table \ref{suptable2}. This demonstrates the highest SR that agent can achieve with explicit awareness of if there are unknown or unseen objects in the current field of view.

\subsubsection{Analysis of Targets}
1): Distance to Target. Unseen objects which does not exist in the training phase make it difficult to establish connections with other known objects, resulting in poor performance in searching phase. Table \ref{suptable3} shows the results when agent has a different distance between the initial positions and targets.
2): Size of Targets. In the unseen environments, the features of small objects are less obvious, making them difficult to be observed and identified. 
In summary, it is crucial to empower agents to obtain comprehensive information pertaining to unseen objects and precisely locate their spatial coordinates. In the future, we will try additional methods for detecting unseen objects and provide the agent with additional location-specific information, such as reference objects for enhancing success rate of navigation.

\section{Conclusion} \label{conclusion}
In our study, we offer valuable insights and effective strategies to address the intricacies of zero-shot object navigation. By incorporating unknown objects into the training phase, we enrich the model with distinct and accessible information. Moreover, we introduce the label-wise meta-correlation module (LWMCM), a novel framework designed to exploit the dynamics between labeled and unlabeled objects, thereby amplifying the depth of object information. Our comprehensive experiments validate the superior performance of our approach, marking a significant advancement in the field of zero-shot object navigation.

\bibliographystyle{IEEEtran}
\bibliography{main}

\end{document}